\pdfoutput=1

\documentclass[11pt]{article}
\usepackage{acl}

\usepackage{times}
\usepackage{latexsym}

\usepackage{color}
\usepackage{amsmath}
\usepackage{amssymb}

\usepackage{CJKutf8}

\usepackage{microtype}
\usepackage{graphicx}
\usepackage{booktabs}
\usepackage{makecell}
\usepackage{multirow}

\title{Towards Automated Real-time Evaluation in Text-based Counseling}

\author{Anqi Li$^{1,2}$, Jingsong Ma$^{2}$, Lizhi Ma$^{2}$, Pengfei Fang$^{2,3}$, Hongliang He$^{1,2}$, Zhenzhong Lan$^{2}$\thanks{\ \ Corresponding author: lanzhenzhong@westlake.edu.cn} \\
$^{1}$ Zhejiang University, $^{2}$ Westlake University, $^{3}$ Australian National University \\
\texttt{\{lianqi, majingsong, malizhi, hehongliang, lanzhenzhong\}@westlake.edu.cn} \\
\texttt{Pengfei.Fang@anu.edu.au} \\
}

\begin{document}
\begin{CJK*}{UTF8}{gbsn}

\maketitle
\begin{abstract}
Automated real-time evaluation of counselor-client interaction is important for ensuring quality counseling but the rules are difficult to articulate. Recent advancements in machine learning methods show the possibility of learning such rules automatically. However, these methods often demand large scale and high quality counseling data, which are difficult to collect. To address this issue, we build an online counseling platform, which allows professional psychotherapists to provide free counseling services to those are in need. In exchange, we collect the counseling transcripts. Within a year of its operation, we manage to get one of the largest set of (675) transcripts of counseling sessions. To further leverage the valuable data we have, we label our dataset using both coarse- and fine-grained labels and use a set of pretraining techniques. In the end, we are able to achieve practically useful accuracy in both labeling system. 
\end{abstract}

\section{Introduction}
\begin{figure}[t]
    \centering
    \includegraphics[scale=0.35]{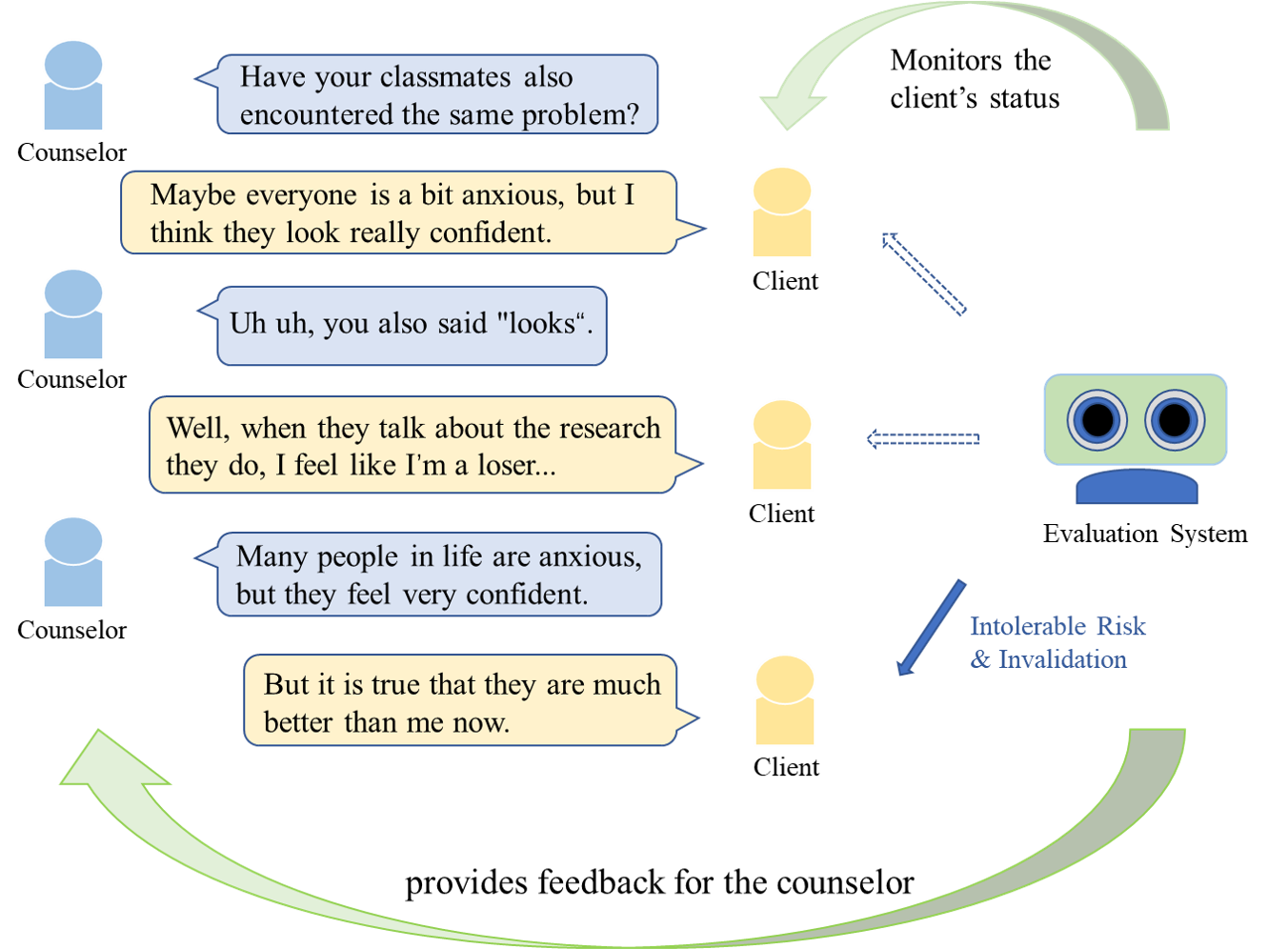}
    \caption{The process of automated real-time evaluation for the psychological counseling in an ongoing session.}
    \label{evaluation_system}
\end{figure}

The last decade has witnessed an increasing demand for psychological counseling in daily life. Although positive effects of psychological counseling are widely recognized~\citep{lambert1994effectiveness, perry1999effectiveness}, it remains great potentiality to improve the quality of counseling services. For instance, the counselors, especially the novice, may provide inappropriate interventions towards clients' maladaptive experience, which would undermine the interaction quality between the counselor and client, leading to a negative outcome of the counseling \citep{1994evaluation}. Therefore, developing an evaluation system that can monitor the quality of each interaction between counselors and clients in an ongoing counseling session is necessary. That is, when the evaluation system detects negative effects in the current interaction, it can provide an immediate feedback to the counselors and suggest them to adjust the perspectives of topics or experiences being discussing in the next turn of counseling dialogue. Figure~\ref{evaluation_system} presents the automated evaluation process, in which the evaluation system identifies the client's status in a timely manner and provides immediate feedback for the counselor in an ongoing session.

Nonetheless, it is not without difficulties to build an evaluation system for psycho-counseling because of some practice issues. 1)-The corpus of psychological counseling is not easy to be accessed due to the confidential nature of psycho-counseling. 2)-It requires the expert-level annotators to label the data. 3)-The advanced machine learning techniques are data-hungry. Adapting machine learning models with few data is also a challenging step. In this paper, we aim to bridge the gap between the machine learning techniques and the evaluation system on psycho-counseling.

To collect the Chinese psycho-counseling corpus in the real-world scenarios, we build an online counseling platform to provide the service of free text-based counselings.

From the platform, we collect 675 transcripts of counseling sessions, conducted by professional counselors employing skills of various counseling genres, including Cognitive Behavioral Therapy (CBT), Narrative Therapy, Psychodynamic Therapy, etc. Taking 3 consecutive counselor-client dialogue turns as a sample, we obtain 33792 samples in total.

To evaluate the quality of the counseling, we further apply the Clients' Response-Experience Types Coding Scheme (CRETCS) to annotate our data. The CRETCS is based on the therapeutic collaboration coding system (TCCS) proposed by \citep{2012How}, and defines the categories of clients' responses and experiences, reflecting the quality of counselor-client interaction from the perspective of clients. Under the principle of the scheme, we can distinguish whether clients accept counselors' interventions and the level of risk that clients experience towards preceding interventions. Following the coding scheme as guidance, we manually annotate 4786 samples among all data.

To study the feasibility of leveraging advanced computational methods to automate the real-time evaluation process, we conduct extensive experiments on the proposed data. To improve the performance of model with limited annotated data on a new domain, we conduct additional pre-training phases to adapt our model to the task-specific data, including the unlabeled data and the annotated training data. The empirical results show that a second pre-training phase is an effective way to bring performance gain. In the detailed error analysis, we find that the model might be confused when categorize two labels that share similar linguistic features.

In this paper, we first construct a Chinese benchmark dataset, which was annotated by a CRETCS. To verify the potential that advanced machine learning-based methods can solve the real-time evaluation of interaction quality in ongoing counseling, extensive experiments were studied on various computational approaches. We believe our benchmark and the proposed evaluation system will bring improvements to the assessment of interaction quality in the field of psycho-counseling. The \textbf{contributions} are summarized as follows:

\begin{itemize}
    \item We first propose a coding scheme for evaluating the quality of the real-time interaction between counselors and clients, and design detailed operational definitions of each category to facilitate our research.
    \item We build a free online text-based psychological counseling platform and collect 675 transcripts of counseling sessions from the platform. We also construct a dataset, which is annotated following the principle of the proposed coding scheme.
    \item Thorough experiments are conducted on our new benchmark dataset. Empirical results indicate that computational approaches hold promise to automate the process of real-time evaluation. 
\end{itemize}

Our code and model will be made available freely to the research community after the paper is accepted. Because of the sensitive and private nature, we will also simulate part of this data and make it available by request.

\section{Related Work}

Our work mainly focuses on the evaluation of psycho-counseling. Here, we briefly give an overview of the assessment methods for the interaction quality in psychological counseling. Thereafter, we also review the computational approaches for the counseling evaluation.  

\subsection{Assessment of the Interaction Quality in Psycho-counseling}

The quality of psychological counseling can be assessed by clients' self-reports of improvements of maladaptive experiences after attending the counseling \citep{1994evaluation}. Meanwhile, the behavioral coding is another primary method employed to evaluate the quality of counseling sessions, in which trained coders assign specific labels or numeric values to each utterance or evaluate the whole sessions based on coding schemes \citep{bakeman2012behavioral}.To evaluate the interaction quality in the counseling, the categorization of each counselors' and clients' utterance during counseling is necessary. The works constructed the coding schemes on the utterance level mainly defined and categorized counselors' strategies, such as support, reflection, question and challenging etc. \citep{lee2019identifying, 2020Using, young1980cognitive}, which reflect therapeutic approaches instead of the interaction quality between counselor and client. Some coding schemes also have taken into account the categories of clients' utterances during counseling but are developed specifically for certain types of counseling in which the categories are closely related to the objectives of the corresponding counseling. For instance, Motivational Interviewing Skill Codes (MISC) was designed to provide the utterance-level labels to code counselors' and clients' behaviors respectively for Motivational Interviewing (MI) \citep{miller2003manual}. While some other studies have designed the coding schemes that can be applied to diverse counseling genres and reflect the interaction between counselor and client. For examples, the categorization of client utterances was developed by \citep{park2019conversation} to represent clients' past experience, current situation and changes of problematic thoughts; and the therapeutic collaboration coding scheme (TCCS) was designed by \citet{2012How} to analyze the effectiveness of moment-by-moment interaction between counselor and client according to clients' zone of proximal development \citep{2001Dialogical}.

\subsection{Computational Approaches for Behavioral Coding of Counseling}

A diverse set of works have adopted advanced computational approaches, which automate the behavioral coding, to evaluate the quality of the text-based counseling~\citep{2012A, xiao2016behavioral, atkins2014scaling, 2016A, gibson2019multi, flemotomos2021automated}. Early works relied mostly on designing hand-crafted feature representations for the text~\cite{2012A, atkins2014scaling, tanana2015recursive, 2016A, perez2017predicting}. One of the earliest work~\citep{2012A} used different sets of hand-crafted linguistic features to automatically identify \emph{Reflection}. In \citep{atkins2014scaling}, the authors adopted topic models to predict counselors' intervention strategies under MISC scheme on the utterance-level. In order to make full use of text information, the work in \citep{perez2017predicting} integrated the syntactic, semantic and similarity features to identify counselors' strategies. In the era of deep learning, a number of solutions employ the advanced neural networks, i.e., convolutional neural network (CNN), recurrent neural network (RNN), Transformer, etc, to extract discriminative features from the text \citep{xiao2016behavioral, huang2018modeling, cao2019observing, lee2019identifying, tavabi2021analysis, flemotomos2021good}.  For example, \citep{cao2019observing} used hierarchical recurrent encoders with attention mechanism to encode dialogues. Some recent works~\cite{2020Using, tavabi2021analysis} benefit from the pre-trained model, including BERT~\citep{devlin2018bert}, RoBERTa~\citep{2019RoBERTa} etc. To detect the categories of clients' responses in counseling of diverse genres, \citep{park2019conversation} develop a pre-trained conversation model using a sequence-to-sequence architecture.

\section{Data Collection}

In this section, we will detail the proposed dataset and the annotation scheme, which can be served as a benchmark to evaluate the counseling quality from the utterances of clients, for various counseling genres. Our dataset is collected from the free Chinese psychological counseling platform, where the counseling services are provided by professional human counselors. In \textsection~\ref{data source}, we first introduce the data selection principles and the statistics of the corpus. Then we will explain the coding scheme CRETCS, which categorizes clients' utterances according to the types of clients' responses and experience in \textsection~\ref{annotation scheme}. Following the coding scheme, we can annotate our data and present a Client Response-Experience Types Dataset (CRETD). The labeling process is discussed in \textsection~\ref{data annotation}. The \textsection~\ref{data analysis} summarizes the characteristics of our dataset.

\subsection{Data Source}
\label{data source}

Our free online Chinese psychological counseling platform provides dyadic text-based counseling for clients. In our online platform, we hire 30 professional counselors, with each counselor experienced with one or more counseling genres, e.g., CBT, Narrative Therapy, Psychodynamic Therapy. Before the time we build our dataset, our platform has served 1405 clients of different genders, ages, levels of education and regions, and the counseling topics include interpersonal relationships, career, self-development, emotion regulation, etc.

Given the counseling transcripts collected from our platform, we choose 675 counseling sessions following two criteria: (1)-the transcripts include more than 20 counselor-client dialogue turns; (2)-the contents in the transcripts reflect an appropriate counseling process, e.g., discussion of maladapted behaviors, undesirable thoughts, difficulties in life, instead of the chit-chat, such as general inquiries about the counseling platform.

Once we obtain the counseling data, we then pre-process each counseling session by following steps: (1)-In each dialogue turn, we combine multiple consecutive messages from the same speaker, connected by a full stop, such that each spliced message is an utterance from the speaker. (2)-We anonymize clients' personal information, including name, organization, etc, for privacy purposes. (3)-In the end, we create our dataset, with each sample consisting of 3 consecutive dialogue turns, started by the counselor. After the pre-processing, we can obtain 33792 unlabeled samples. In Table~\ref{sample snippet}, we show a sample snippet in both Chinese and its English counterpart. Statistically, the average dialogue turns per session are 50 in the counseling data, and in the pre-processed samples, the average length of counselors' and clients' utterance is 21 and 32 characters respectively. The label distribution of annotated data in shown in Table \ref{data distribution}.

\begin{table*}[!ht]
    \centering
    \scalebox{0.78}{
    \begin{tabular}{c|c|c|c|c}
        \toprule
        Speaker &  Utterance & Index & Response Type &  Experience Type\\
        \midrule
        ... & ... & ...& -- & --  \\
        \midrule
        counselor & \textcolor{blue}{\makecell[l]{听起来像是，本质上不用承担选择的后果和责任。 \\ It sounds like, essentially, you don’t have to bear the consequences \\ and responsibilities of your choice.}} & $t_1$ & -- & -- \\ 
        \midrule
        client & \textcolor{blue}{\makecell[l]{可是这个被安排的结果，也还是我自己承担嘛，我是觉得差不太多。 \\ But I still have to bear the result of this arrangement. I don't think \\ the difference is too much.}} & $c_1$ & Invalidation & Intolerable Risk \\
        \midrule
        counselor & \textcolor{blue}{\makecell[l]{免去了自己选择别的可能性的后果 \\ But it could eliminate the consequences of choosing other possibilities.}} & $t_2$ & -- & -- \\ 
        \midrule
        client & \textcolor{blue}{\makecell[l]{确实。 \\ Exactly.}} & $c_2$ & Validation & Safety  \\
        \midrule
        counselor & \textcolor{orange}{\makecell[l]{那对于没有被安排的其他部分，你怎么看呢？ \\ What do you think about the other parts that are not arranged?}} & $t_3$ & -- & -- \\
        \midrule
        client & \textcolor{purple}{\makecell[l]{因为面对的选择太多，结果也不确定，显而易见，我选择了逃避，\\ 这也让我更糟糕了。 \\ Because there are too many choices, the result is not certain. \\ Obviously, I chose to escape, which made me worse.}} & $c_3$ & Validation & Tolerable Risk \\
        \midrule
        ... & ... & ...& -- & -- \\
        \bottomrule
    \end{tabular}}
    \caption{A sample snippet of a simulated transcript with its English translated version. Each sample is composed of three consecutive counselor-client dialogue turns. Each \textcolor{purple}{client's utterance} is annotated with response and experience type towards the \textcolor{orange}{counselor's utterance} with taking \textcolor{blue}{dialogue history} as context.}
    \label{sample snippet}
\end{table*}

\subsection{Annotation Scheme}
\label{annotation scheme}

As mentioned above, coding the counselors' utterances reflects only the therapeutic approaches rather than the interaction quality between counselor and client during counseling ~\citep{2020Balancing}. As an example shown in Figure~\ref{evaluation_system}, the client's refusal to counselor's intervention, who tried to change the client's problematic thoughts, indicating that the counselor's utterances might not be a suitable intervention as to the client at that moment. However, this utterance of the counselor may be evaluated as an ``appropriate" intervention based on the therapeutic techniques solely \citep{1987Some, 1983Cross, 2020Balancing}. Thus, it is necessary to consider the clients' responses when assessing the interaction quality during ongoing counseling.

To address the issue, our idea is to evaluate the effectiveness of counselors' intervention and the interaction quality by means of analyzing the types of response and experience expressed by the clients.

\begin{figure}[t]
    \centering
    \includegraphics[scale=0.45]{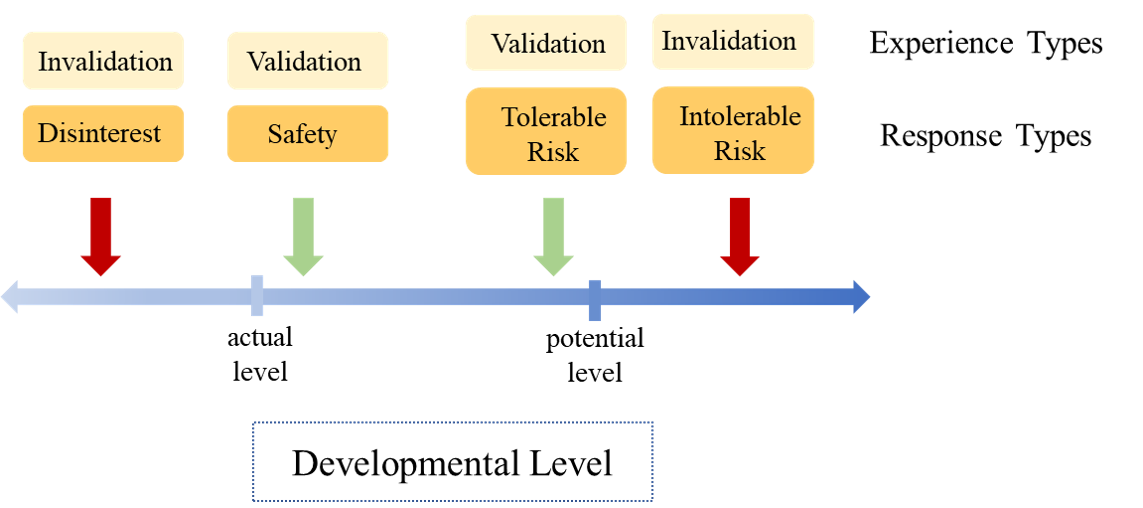}
    \vspace{-3mm}
    \caption{The relationship among \textcolor{blue}{clients' developmental level}, \textcolor{red}{counselors'  intervention}, \textcolor{orange}{clients' experience} and \textcolor{yellow}{clients' response types}.}
    \label{fig:TZPD}
\end{figure}

{Referring to Vygosky's zone of proximal development \citep{vygotsky1978development}, the clients attending the psycho-counseling develop their problem solving skills within the development level and experience different degrees of risk in response to counselors' interventions (see Figure ~\ref{fig:TZPD}) \citep{2001Dialogical}. If counselors' interventions inspire clients to explore and solve their problems effectively, the clients will follow the interventions within their developmental level (between the \emph{actual} and \emph{potential level}). They might feel safe to communicate with the counselor or feel challenging when discussing their maladaptive experience, indicating that the clients are making valid responses. But if the counselors' interventions fail to capture the clients' attention (below the \emph{actual level}) or irritate the clients (above the \emph{potential level}), the clients might lose interests in the no-going conversation or start to defend themselves, reflecting that their responses become invalid as to counselors' interventions. Based on these concepts, in the CRETCS, we categorize clients' utterances into \emph{Validation}, \emph{Invalidation} as to the response types, and \emph{Safety}, \emph{Tolerable Risk}, \emph{Intolerable Risk} and \emph{Disinterest} as to the experience types, which is adapted from the categorization of clients' responses in TCCS~\citep{2012How}. Additionally, we add the label \emph{Others} to code utterances that counselors and clients are not talking about the clients' problems, including greetings, scheduling the next appointment and expressing appreciations to counselors etc.}

Considering the fact that annotators' subjective perception on distinguishing "the level of risk" from clients' experience may affect the accuracy of annotation, we designed a handbook, which contains the detailed and objective operational definitions of all the categories and the corresponding examples, to facilitate the annotation process. The handbook was finalized through repeated procedures of annotation, discussion of resolving differences and reaching consensus, analysis of reasons, and summary of criteria by two research psychologist, one of them holds the certificate of psycho-counseling qualified by Ministry of Human Resource and Social Security of China. In the revision process, the research psychologists annotated three sessions with 158 samples. 42 out of the samples are used as examples in the handbook and the remaining was used to train and test the labeling accuracy of annotators before formally annotation (detailed in \textsection~\ref{data annotation}). Samples of the handbook will be shown in the { Appendix~\ref{handbook}}.

\subsection{Data Annotation}
\label{data annotation}

Having the utterance samples and coding scheme at our disposal, we are ready to annotate our data. 
As shown in Table~\ref{sample snippet}, each sample includes three dialogue turns and can be split into dialogue history (the \textcolor{blue}{blue} text), counselor's intervention utterance (the \textcolor{orange}{orange} text), client's response utterance (the \textcolor{purple}{purple} text). Considering the context from dialogue history and counselors' intervention, we can annotate first the \textit{response type} and then \textit{experience type} w.r.t. client's response utterance according to the coding scheme.

Accurate annotation is another challenging task to build our psycho-counseling dataset. In doing so, we hired and trained 5 graduate students majoring in Psychology to annotate the samples. Before the formal annotation, all the students were required to read the handbook about the definitions and examples. To evaluate their understandings, an examination was adopted to test the labeling accuracy of the students. Such a process was repeated and two students, whose mark of the final exam is above 80$\%$, are selected as the final annotators. In the formal procedure of annotation, two annotators annotated the same data independently. The samples annotated with different labels from two annotators would be revised by the research psychologist with the certificate of counseling. Finally, our dataset includes 4786 samples.

\subsection{Data Analysis} \label{data analysis}

Our dataset has two characteristics:

{\noindent\textbf{Hierarchy of labels.} Each utterance sample is annotated by two types of labels, i.e., response type and experience type. These two types of labels follow the hierarchical relationship of two scales -- the response type is the coarse-level annotation while the experience type is the fine-level annotation. That said, the same response type may result from different experience types.}

{For example, both the experience of \emph{Intolerable Risk} and \emph{Disinterest} leads to the \emph{Invalidation} towards counselors' interventions.}

\noindent\textbf{Imbalanced data distribution.} To better understand the dataset, we illustrate the statistics of the proposed dataset in Table ~\ref{data distribution}. As shown in Table~\ref{data distribution}, the number of \emph{Validation} is about five times larger than that of \emph{Invalidation} (i.e., 3553 vs. 559) in the coarse-level annotation. While in the fine-level annotation, the ratio of \emph{Safety} to \emph{Tolerable Risk} is about 1.6~$:$~1 and the number of \emph{Intolerable Risk} and \emph{Disinterest} are very similar. Such a data distribution is inline with the fact \citep{1994evaluation, ackerman2003review} that counselors spent most of time to intervening gently to support the clients and make them feel safe to form a positive alliance, and then challenge clients to face their difficulties independently within their developmental level. But the counselors also inevitably provide inappropriate interventions occasionally making the clients lose interest or become defensive during counseling. While the label \emph{Others}, which is used to code the utterances of greetings, appreciation or schedule of appointment, generally only appears at the beginning and end of each session.

We also present the mean length of clients' utterances of each category in Table~\ref{data distribution}. We notice that there is no significant difference in the average length of clients utterances when they express \emph{Validation} or \emph{Invalidation} responses, while the mean length of \emph{Others} is distinctly shorter than other types. As for experience types, the mean length of clients' utterances labeled with \emph{Disinterest} and \emph{Tolerable Risk} is much longer than that of \emph{Safety} and \emph{Intolerable Risk}, which is because clients tend to provide more information to counselors when they experience \emph{Tolerable Risk} or talk in a wordy manner to explain a non-meaningful story when they experience \emph{Disinterest}.

\begin{table}[]
    \centering
    \scalebox{0.57}{
    \begin{tabular}{|c|c|c||c|c|c|}
        \hline
        {Response Type} & {Num} & {Mean Length} & {Experience Type} & {Num} &{Mean Length}\\
        \hline
        \multirow{2}{*}{Validation} & \multirow{2}{*}{3553} &\multirow{2}{*}{29.83} & Safety & 2174 & 24.45\\
        & & & Tolerable Risk  & 1379 & 38.30 \\
       \hline
        \multirow{2}{*}{Invalidation} & \multirow{2}{*}{559} & \multirow{2}{*}{30.68} & Intolerable Risk & 326 & 27.86\\
        & & & Disinterest & 233 & 34.62 \\
        \hline
        Others & 674 & 8.81 & Others & 674 & 8.81\\
        \hline
    \end{tabular}
    }
    \caption{Data distribution and mean length of each category under response types and experience types, respectively.}
    \label{data distribution}
\end{table}

\section{Experiments}

In this section, extensive experiments are conducted on the proposed dataset for both fine- and coarse-grained classification. Considering its hierarchical property, we also design a set of models to evaluate the dataset.

\subsection{Data Preparation}

We first define the standard training protocol of the annotated dataset. The dataset is randomly split into training set (70$\%$), validation set (15$\%$) and test set (15$\%$). Noted in the split, the proportion per label of each set keeps unchanged. The corpus of the unlabeled data is randomly split into training set (90$\%$) and validation set (10$\%$), will be used for task-adaptive pre-training task.

\subsection{Network Architectures}
\label{network architectures}
\paragraph{Baseline}

We use the pre-trained Chinese  BERT base model \citep{devlin2018bert}, provided by HuggingFace~\footnote{https://github.com/huggingface}, as our baseline for both the coarse- and fine-grained classification tasks. Following the common practice~\citep{devlin2018bert}, we add a feed-forward layer on top of the [CLS] token as a task-specific classifier, where [CLS] is the classification token in BERT.

\subsection{Training Methods}
\label{training methods}

\paragraph{Task-adaptive Pre-training.} In order to attain a better model with limited annotated data on the domain of psychological counseling, we conduct a second phase of pre-training to adapt our model to the task-specific data, whose effects of gaining performance have been demonstrated in \citep{chronopoulou2019embarrassingly, 2020Don}. We refer to this method as task-adaptive pre-training or TAPT for short. Given the fact that we have a small size of training data and a larger pool of unlabeled data, we set two kinds of task-adaptive pre-training experiments.

\noindent\textbf{TAPT.} In this experiment, we use an auxiliary masked language model (MLM) loss during training in the fine-tuning process. The final loss function is the weighted sum of the task-specific classification loss and the auxiliary MLM loss.

\noindent\textbf{Curated-TAPT.} In this pipeline, we first pre-train a BERT model on the unlabeled data with an MLM loss. During training, we select the best checkpoint with the minimal MLM loss value on the validation set. Then we perform fine-tuning on the small set of annotated data on the model of the best checkpoint. Noted in the fine-turning stage, the MLM loss is not used in this experiment.

\noindent\textbf{Curated-TAPT+TAPT.} In this setting, we develop a pipeline combining the process of `TAPT' and `Curated-TAPT'. That said, the Curated-TAPT is used for pre-training of the model on the task-specific unlabeled data, followed by TAPT for fine-turning on the annotated training data.

\subsection{Implementation Details}
\paragraph{Model Input.}

To identify speaker information in counseling, we use [T] and [C] as special tokens to represent the utterance for counselor and client respectively, whose embeddings will also be trained in the pre-training and fine-tuning process. For each sample with the format of (dialogue history, counselor's intervention utterance, client's response utterance) as shown in Table~\ref{sample snippet}, we add a special token ([T] or [C]) at the beginning of each speaker's utterance and concatenate all utterances into a flat sequence with a separation token ([SEP]) before the client' response utterance. Taking the sample in  Table~\ref{sample snippet} as an example, the data is presented as a sequence of $(t_1, c_1, t_2, c_2, t_3, c_3)$, where $t_i$ and $c_i$ are the counselor and client's utterance of the $i$-th dialogue turn. In the pre-training stage, the sample is fed in the form of ``[CLS][T]$t_1$[C]$c_1$[T]$t_2$[C]$c_2$[T]$t_3$[SEP][C]$c_3$" with some non-special tokens are masked. While in the fine-tuning stage, the input sample is presented as:``[CLS][T]$t_1$[C]$c_1$[T]$t_2$[C]$c_2$[T]$t_3$[SEP][C]$c_3$".

\paragraph{Experimental Settings.} 
All the models are implemented with PyTorch deep learning package~\citep{2019PyTorch}. In pre-training process, the masking probability in the MLM task is set to 0.15. In the fine-turning stage, we initialize weights of feed-forward layers with normal distribution. We select the model that achieves best macro-F1 value on the validation set to test on the test set. For all training process, we adopt cross-entropy loss as the default classification loss. And we use Adam optimizer to train the network with momentum values {$[\beta_1, \beta_2] = [0.9 , 0.999]$}. The learning rate is initialized to $5e-5$ and decayed by using the linear scheduler. The batch size in the training stage is 8. All our experiments are performed on one NVIDIA A100 GPU.

\subsection{Experimental Results}

Extensive experiments are conducted on the baseline, with each being evaluated with three different pre-training strategies (i.e., TAPT,  Curated-TAPT, Curated-TAPT+TAPT). We follow the standard protocol to evaluate the models with `accuracy', `precision', `recall', and `macro-F1' metrics. The results for coarse- and fine-grained classification are illustrated in Table~\ref{tab:experimental results of coarse-grained} and Table~\ref{tab:experimental results}, respectively. As compared with the results of coarse-grained classification, we find the fine-grained classification is very challenging. More detailed results about precision and recall for each category are included in the appendix \ref{appendix:each category results}.

We find that: along with the vanilla pre-trained BERT, additional pre-training on task-specific data, like TAPT and Curated-TAPT, can consistently bring performance gain for both-level classification tasks. Another interesting observation is that the TAPT always performs better than the Curated-TAPT. Moreover, applying TAPT after adapting to the extra task-specific unlabeled data (i.e., Curated-TAPT+TAPT) leads to better performance and this pipeline of pre-training brings more improvements on coarse-grained than fine-grained classification compared to each baseline. Specifically, taking the baseline model as a reference in coarse-grain classification, Curated-TAPT brings an improvement of 2.9$\%$ on macro-F1, while applying TAPT results in a big boost in performance by nearly 8.2$\%$. As compared to TAPT, we also demonstrate that Cuarated-TAPT + TAPT can further improve the result by nearly 1$\%$ and attain the best performance.

\begin{table*}[ht]
    \centering
    \scalebox{0.8}{
    \begin{tabular}{c|ccc|cccc}
        \toprule
        \multirow{2}{*}{\textbf{Model Architectures}} & \multicolumn{3}{c|}{\textbf{Additional Pretraining Phases}} & \multirow{2}{*}{\textbf{Acc.}} & \multirow{2}{*}{\textbf{Precision}} & \multirow{2}{*}{\textbf{Recall}} & \multirow{2}{*}{\textbf{Macro F1}} \\
        & Curated-TAPT &  TAPT & Curated-TAPT + TAPT & & & & \\
        \hline
        \multirow{4}{*}{Baseline} & & & & $83.79_{1.01}$ & $71.59_{1.48}$ & $69.41_{1.89}$ & $69.91_{1.68}$ \\
        \cline{2-8}
         & \checkmark & & & $85.54_{0.73}$ & $75.36_{2.15}$ & $70.93_{2.77}$ & $71.93_{2.17}$ \\
         & & \checkmark & & $86.14_{1.19}$ & $77.54_{2.07}$ & $74.78_{1.95}$ & $75.65_{1.28}$ \\
         & & & \checkmark & $\boldsymbol{86.42_{1.44}}$ & $\boldsymbol{78.76_{2.09}}$ & $\boldsymbol{75.20_{2.64}}$ & {$\boldsymbol{76.27_{1.47}}$}\\
        \bottomrule
    \end{tabular}}
    \caption{The comparative results on the coarse-grained categories of all models. We report averages across ten random seeds, with standard deviations as subscripts. Best task performance is boldfaced.}
    \label{tab:experimental results of coarse-grained}
\end{table*}

\begin{table*}[ht]
    \centering
    \scalebox{0.8}{
    \begin{tabular}{c|ccc|cccc}
        \toprule
        \multirow{2}{*}{\textbf{Model Architectures}} & \multicolumn{3}{c|}{\textbf{Additional Pretraining Phases}} & \multirow{2}{*}{\textbf{Acc.}} & \multirow{2}{*}{\textbf{Precision}} & \multirow{2}{*}{\textbf{Recall}} & \multirow{2}{*}{\textbf{Macro F1}} \\
        & Curated-TAPT & TAPT & Curated-TAPT + TAPT & & & & \\
        \hline
        \multirow{4}{*}{Baseline} & & & & $68.25_{1.95}$ & $59.54_{3.36}$ & $54.91_{1.82}$ & $55.81_{2.19}$ \\
        \cline{2-8}
         & \checkmark & & & $67.73_{1.27}$ & $59.30_{1.50}$ & $55.59_{1.36}$ & $56.67_{1.26}$ \\
         & & \checkmark & & $68.98_{1.65}$ & $\boldsymbol{62.26_{2.49}}$ & $57.53_{2.13}$ & $58.49_{2.23}$ \\
         & & & \checkmark & $\boldsymbol{69.08_{0.87}}$ & ${61.98_{1.60}}$ & {$\boldsymbol{59.08_{1.32}}$} & {$\boldsymbol{59.72_{0.70}}$} \\
        \bottomrule
    \end{tabular}}
    \caption{The comparative results on the fine-grained categories of all models. We report averages across ten random seeds, with standard deviations as subscripts. Best task performance is boldfaced.}
    \label{tab:experimental results}
\end{table*}

\subsection{Error Analysis}
In this section, we analyze the performance of the best-performing model for coarse- and fine-grained classification respectively.

\noindent\textbf{Coarse-grained categorization.} The confusion matrix for the categorization of response and experience types is shown in Appendix~\ref{confusion matrix}. The overall performance of the coarse-grained model is limited by \emph{Invalidation} with 0.51 F1-score (refer to {Appendix \ref{appendix:each category results}} for detailed results). To better understand the confusion between \emph{Invalidation} and \emph{Validation}, we manually analyze 46 \emph{Invalidation} samples that are predicted as \emph{Validation} and 27 samples vice versa. In this process, we find that the most typical cases are that the model categorizes utterances containing negative words, especially "not" or "no", into \emph{Invalidation} because many samples in the \emph{Invalidation} include these words. For instance, "\textbf{\textit{Counselor: I think you hold a positive attitude towards the future. Client: Yes, you are right. In fact, the overall condition of me is not bad, but I still cannot maintain a close relationship}}". Likewise, the majority of samples in \emph{Validation} contains "yes" and "ok", the model tends to categorize clients' utterances into \emph{Validation} when "yes" or "ok" is detected. But the utterances are manually categorized as \emph{Invalidation} because the clients fails to respond to counselors' intervention: "\textbf{\textit{Counselor: You've already known that the people you met and what happened in your life does not help you to achieve your goal. Then, what's your plan? Client: Yes. I'm so tried, coping with all these stuff. It may still end up with nothing.}}" Thus, the performance of model might be affected by the shared semantic meaning of common words across the \emph{Validation} and \emph{Invalidation}.

\noindent\textbf{Fine-grained categorization.} The confusion matrix of the fine-grained classification is presented in Appendix~\ref{confusion matrix}. The primary error of the fine-grained model comes from label \emph{Disinterest} with only 0.26 F1-score and \emph{Intolerable Risk} with 0.47 F1-score, which is largely caused by data imbalance. In the confusion matrix, some \emph{Intolerable Risk} samples are predicted as \emph{Tolerable Risk}, which is reasonable due to the limited dialogue history put into BERT which is useful to distinguish \emph{Intolerable Risk} from \emph{Tolerable Risk}. Meanwhile, the model confuses the label \emph{Safety} and \emph{Tolerable Risk}, which is partly due to the model lacks knowledge of counselors' intervention strategies. For instance, the counselors ask questions to, first, get demographic or experience-related information from clients where the clients' answers are labeled as \emph{Safety}. Second, the counselors' questions are to further explore or provide new perspective for clients to cope with their existing problems where the proper answers are labeled as \emph{Tolerable Risk}. Specifically, the case "\textbf{\textit{Counselor: Then, what game is it? Client: [Name of the game], my friends recommended to me, I do like this type of games}}" is \emph{Safety} but predicted as \emph{Tolerable Risk}; the case "\textbf{\textit{Counselor: In this situation, how did you cope with your anxiety? Client: I found some other things to do to ease the anxiety. But after doing so, I became more anxious because I had less time to deal with the stuff I should have done.}}" is \emph{Tolerable Risk} but was labeled as \emph{Safety}. As a result, without knowing the intervention strategies, the model might be confused when categorizing the samples of \emph{Tolerable Risk} and \emph{Safety} due to the similar structures of counselor-client interaction.

In addition, \emph{Others} is the easiest to categorize in both classification tasks as expected. Because samples coded as \emph{Others} are quite different from other samples in terms of semantics and length.

\section{Conclusion}
In this work, we aim to address the issue of real-time evaluation of psychological counseling with various genres. Considering to avoid inappropriate interventions from the counselors, we adopt a new coding scheme (e.g., TCCS) to annotate the clients' utterances in our new benchmark dataset. We also study the property of the proposed dataset via various advanced deep learning methods. 

In future work, we will develop more advanced machine learning solutions to improve the performance of the evaluation systems.

\section*{Ethical Considerations}
In the proposed dataset, we have omitted the personal information for the counselors and clients to protect their privacy and identity. It is also not the case that our model can be used to evaluate all kind of psychological counseling for reasons that every dataset is subject to its intrinsic bias and the trained models will inevitably learn the biased characteristics in the dataset.

\bibliography{custom}
\bibliographystyle{acl_natbib}

\clearpage
\appendix

\section{Appendix}

\subsection{Definitions of Categories in Clients' Response-experience Type Coding Scheme}
\label{handbook}

The categories of clients' utterances, response types (\emph{validation} and \emph{invalidation}) and experience types (\emph{safety}, \emph{tolerable risk}, \emph{disinterest}, \emph{intolerable risk}) are depicted below. The definitions of each category are adapted from the therapeutic collaboration coding scheme (TCCS) \citep{2012How}. Meanwhile, examples with English translation of counselor-client interactions, simulated from the original data, are provided to facilitate the understanding of our coding process. 

\subsubsection*{------------------Validation -- Safety-----------------}

\textbf{\textit{a. Client agrees with counselor’s intervention but does not expand it when the counselor describes or summarizes the client’s discourse using his/her own or client’s words.}}

咨询师：这些活动给你带来了成就感，但是占用太多时间，对学习和休息都有影响。

来访者：是的

Counselor: These activities brought you a sense of achievement, but took up too much time and affected your study and rest.

Client: Yes.

\noindent\textbf{\textit{b. Client answers counselor's questions when the counselor explores clients’ feelings, ideas and experience using open questions.}}

咨询师：他们给你介绍男朋友的时候，你是什么反应呢？

来访者：要么告诉他们在谈着，要么就说现在还不想找。

Counselor: How did you react when they wish to introduce you to potential boyfriends?

Client: I would tell them that I had already had one or I don’ t want to make a boyfriend now.

\noindent\textbf{\textit{c. Client seeks advice, explanations, suggestions from the counselors.}}

来访者：很希望你可以教教我该怎么更好地去应对。

Counselor: I wish you could tell me how to deal with it in a better way.

\subsubsection*{------------Validation -- Tolerable Risk------------}

\textbf{\textit{a. Client accepts the counselor's advice/suggestions when the counselor invites the client to act in a different way during or after the counseling.}}

咨询师：你有尝试走出去，去别的地方学习吗？

来访者：还没有，我想可以尝试一下。

Counselor: Have you ever tried to go outside and to study elsewhere?

Client: Not yet. But I think I could give it a go.

\noindent\textbf{\textit{b. Client reformulates his or her perspective over the experience being explored when the counselor directly or implicitly points out the client's maladaptive beliefs or behaviors.}}

咨询师：觉得麻烦到他，其实只是你自己的想法。

来访者：或许真的只是我自己单方面在忧虑这些。

Counselor: In fact, it is just your own thinking that this person was bothered by you.

Client: Perhaps it's true, it is only me who is worrying about these stuff.

\noindent\textbf{\textit{c. Client attempts to clarify the sense of or make further explanation to his or her previous statement to the counselor.}}

来访者：感觉很无助，很委屈。

咨询师：觉得自己没有被理解。

来访者：嗯嗯，就像是被关在一个笼子里怎么也出不去。

Client: I feel really helpless and wronged.

Counselor: You think that you are not understood by others. 

Client: Yes, it kinds of like that I was being locked in a cage and could not get out at all.

\subsubsection*{----------Invalidation -- Intolerable Risk----------}

\textbf{\textit{a. Client feels confused and/or are unable to answer the counselor’s question when the counselor points out the client's maladaptive beliefs, feelings or behaviors by asking questions or speaking out directly.}}

咨询师：你说那件事对你好像没有影响，但是你依然在意，这是因为什么呢？

来访者：想不到原因。

Counselor: You have said that it seems like you were not affected by that thing. But you still care about it. Why?

Client: I have no idea.

\noindent\textbf{\textit{b. Client does not accept counselor's advice or suggestions, but persist on looking at their specific experience or topic from his or her perspective when the counselor implicitly or explicitly invites the client to look at the given experience or topic in an alternative way.}}

咨询师：我看到了“敌人”对你的影响，它对你的帮助和限制是什么？

来访者：我被打倒了，没有什么帮助。它使我失去了很多战胜困难的机会。

Counselor: I understand the influence of "Enemy" on you. How does it help and limit you?

Client: I am defeated by it, there isn't any helps. It makes me lose so many opportunities to overcome difficulties. 

\noindent\textbf{\textit{c. Client refuses to accept counselor's advice or suggestion when the counselor implicitly or explicitly invites the client to act in a different way during or after the counseling.}}

咨询师：听起来妈妈的确不觉得这件事情对你来说很重要。你是否尝试过“事件＋感受”的表达方式。例如，妈妈这样做会让我觉得难堪，如果妈妈能为我保密我会很高兴。

来访者：我不想试。

Counselor: It sounds like that your mother did do not think the thing was important to you. Have you ever tried to express your feelings in a way of “events plus feelings”? Like, “Mum, what you did does embarrass me”, “Mum, I would be happy if you kept the secret for me.”.

Client: I don't bother to try.

\subsubsection*{-----------Invalidation -- Disinterest [ID]----------}

\textbf{\textit{a. Client changes topic or peripherally answers the counselor's intervention.}}

咨询师：现在一切事情都在往好的方面发展啊。

来访者：我想去另一个城市。

Counselor: Everything is getting better and better now.

Client: I want to go to another city.

\noindent\textbf{\textit{b. Client talks in a wordy manner or overly elaborates non-significant stories to explain an experience that is not related with counselor’s intervention.}}

咨询师：我能感受到你的无力感以及愤怒，是什么阻碍了你去改变呢？

来访者：上级让他告诉我去一个地方，他没有给我打电话，却跟上级

说给我打电话我不接。他还说我的坏话...

Counselor: I can feel your powerlessness and anger. But what keeps you from changing?

Client: The boss asked him to tell me to come to the place. In fact, he did not call me, but he told the boss that I did not answer the phone. He often speaks ill of me...

\noindent\textbf{\textit{c. Client questions counselor’s intervention or replies in a sarcastic manner.}}

咨询师：你随时可以离开他们也没关系。

来访者：我感觉目前聊下来，好像没什么太多的帮助…

Counselor: It would be fine, you could leave them for elsewhere anytime.

Client: So far, I feel our conversation is not helping.

\noindent\textbf{\textit{d. Client gives little responses to counselor’s intervention and/or is reluctant to think about what the counselor wish to explore.}}

咨询师：你觉得她找你是为了什么呢？

来访者：这我不知道，我也不想知道，我不在乎这个。

Counselor: What do you think the reason why she comes for you?

Client: Well, I don't know and don't want to know. I don't care about it.

\subsubsection*{---------------------------Others------------------------}

\textbf{\textit{a. Greeting each other at the beginning of the counseling.}}

咨询师：你好！

来访者：你好！

Counselor: Hello!

Client: Hi!

\noindent\textbf{\textit{ b. Client gives feedback at the end of the counseling.}}

咨询师：今天聊下来感觉如何？

来访者：挺好的。

Counselor: How do you feel now after the whole conversation today?

Client: Pretty well.

\noindent\textbf{\textit{c. Schedule the next counseling.}}

咨询师：下次咨询什么时间比较方便呢？

来访者：下周一19:00是可以的。

Counseling: What time is suitable for the next counseling?

Client: Next Monday at 19:00.

\noindent\textbf{\textit{d. Client shows appreciations to the counselor.}}

咨询师：我们下周接着聊。

来访者：好的，谢谢你！

Counselor: I will see you next week.

Client: OK, Thank you!

\subsection{Experimental Results for Each Category}
\label{appendix:each category results}
Table~\ref{tab:coarse} and Table~\ref{tab:fine} show the detailed experimental results of each category of the coarse- and fine-grained classification of the best-performed model respectively.

\begin{table}[]
    \centering
    \begin{tabular}{c|ccc}
        \toprule
        \textbf{Category} & \textbf{Prec.} & \textbf{Recall} & \textbf{F1} \\
        \hline
        Validation &  $0.91$ & $0.93$ & $0.92$ \\
        Invalidation & $0.58$ & $0.45$ & $0.51$\\
        Others & $0.91$ & $0.94$ & $0.93$ \\
        \bottomrule
        
    \end{tabular}
    \caption{The coarse-grained classification results of each category.}
    \label{tab:coarse}
\end{table}

\begin{table}[]
    \centering
    \begin{tabular}{c|ccc}
        \toprule
        \textbf{Category} & \textbf{Prec.} & \textbf{Recall} & \textbf{F1} \\
        \hline
        Safety & $0.72$ & $0.73$ & $0.73$  \\
        Tolerable Risk & $0.63$ & $0.70$ & $0.66$ \\
        Intolerable Risk & $0.61$ & $0.39$ & $0.47$ \\
        Disinterest & $0.31$ & $0.23$ & $0.26$ \\
        Others & $0.92$ & $0.93$ & $0.93$ \\
        \bottomrule
    \end{tabular}
    \caption{The fine-grained classification results of each category.}
    \label{tab:fine}
\end{table}

\subsection{Confusion Matrix}
\label{confusion matrix}
\begin{figure}[!ht]
    \centering
    \includegraphics[scale=0.12]{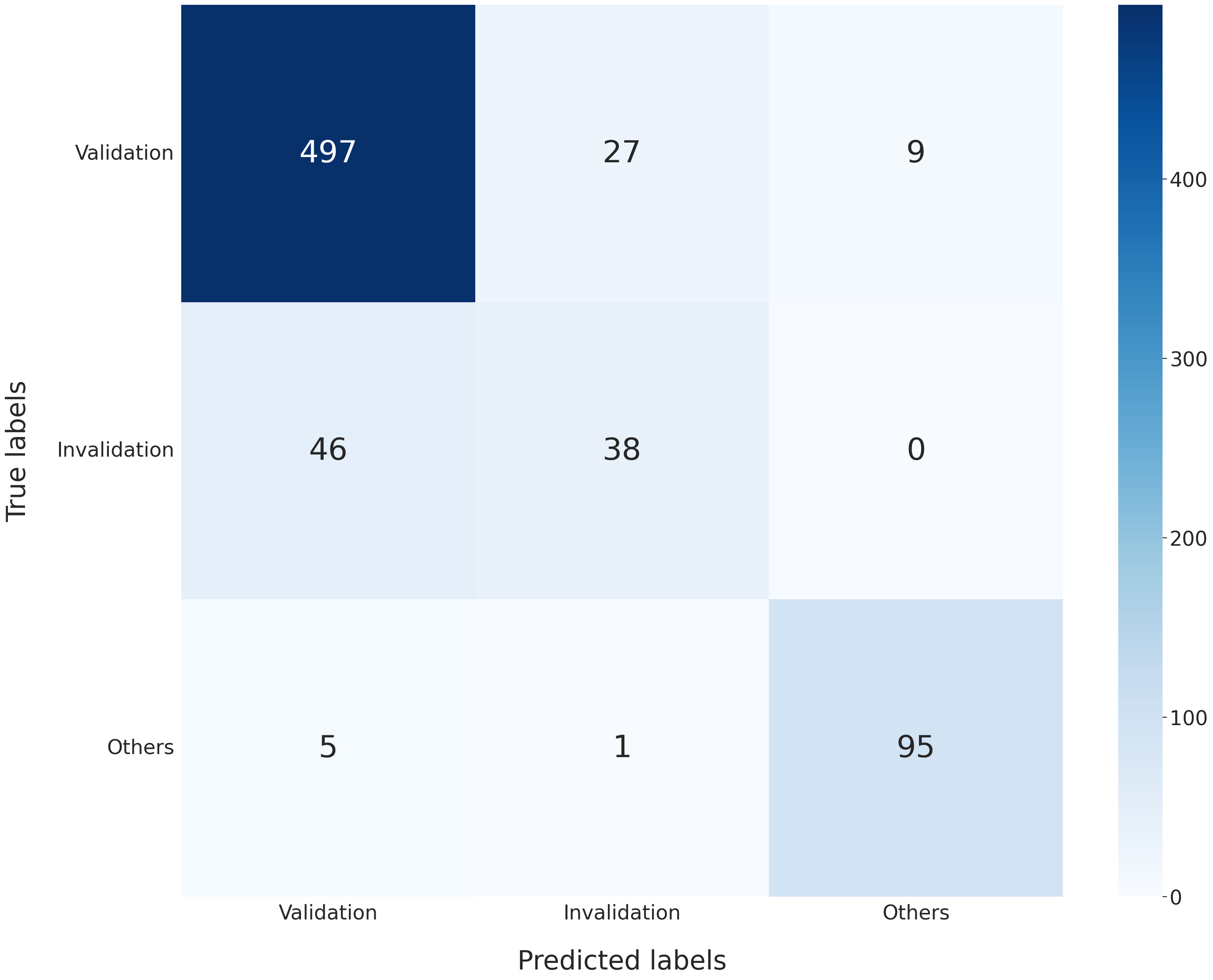}
    \caption{Confusion matrix for coarse-grained classification}
    \label{fig:coarse-grained confusion matrix}
\end{figure}

\begin{figure}[!ht]
    \centering
    \includegraphics[scale=0.1]{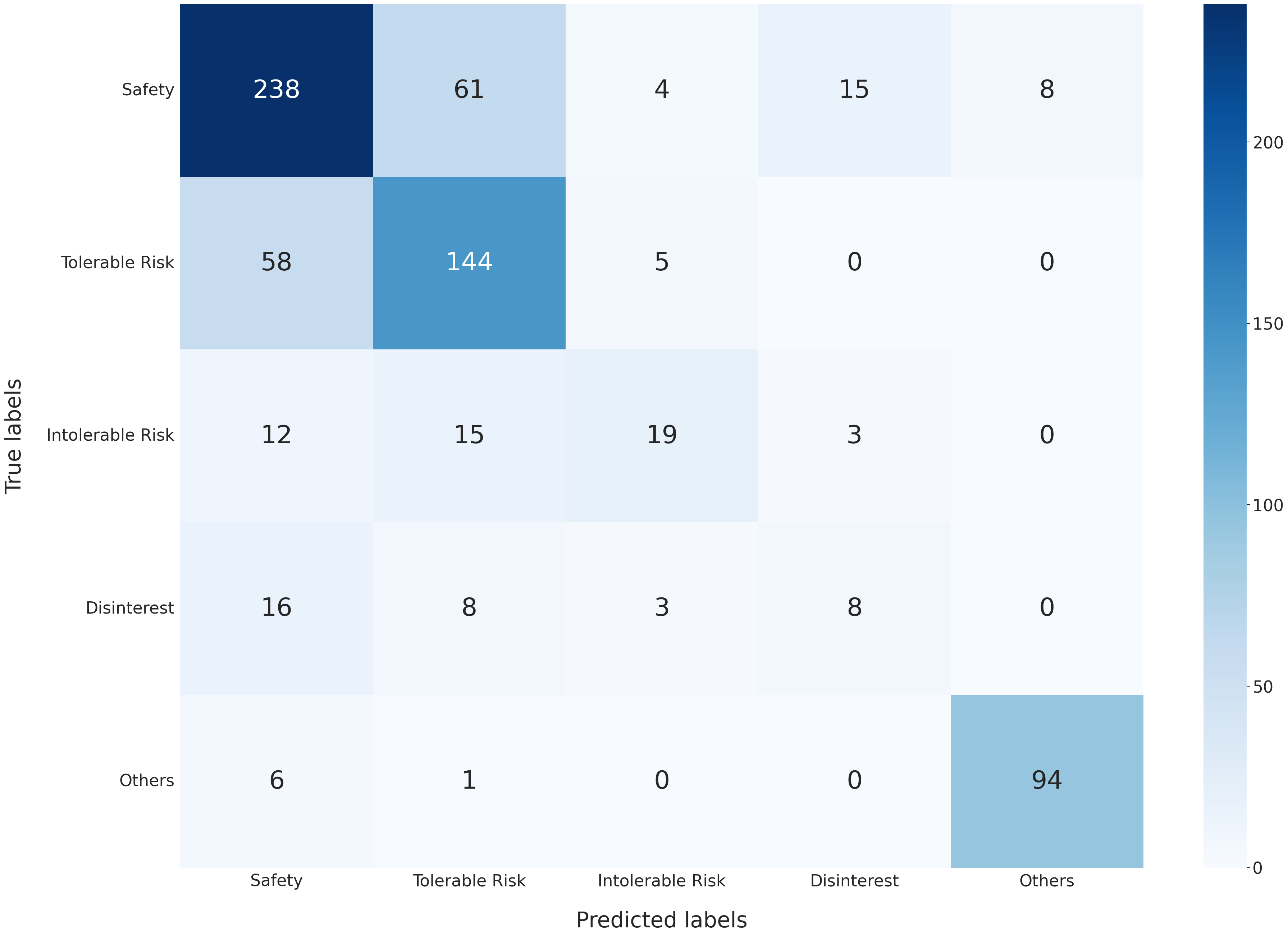}
    \caption{Confusion matrix for fine-grained classification}
    \label{fig:fine-grained confusion matrix}
\end{figure}

Figure~\ref{fig:coarse-grained confusion matrix} and Figure~\ref{fig:fine-grained confusion matrix} show the confusion matrix for the coarse- and fine-grained classification of the best-performed model respectively.

\subsection{The Consent Form of Online-counseling}
\label{consent form}

Below is the consent form used in the current work. Every client gave their consent to attend the online text-based psycho-counseling on our counseling platform and agreed to data usage for the current work. 

0: 很高兴能为您服务。

0: I am very happy we can serve you.

1: 为了保障您的权益，请在开启对话之前 阅读并同意使用须知

1: In order to protect your rights, please read and agree to the instructions before starting the conversation

2: 我同意接受此心理咨询平台根据我的困惑提供线上文字咨询服务（以下简称服务）

2: I agree to accept the current counseling platform to provide online text-based counseling services based on my confusion (hereinafter referred to as the service)

3: 我了解现阶段此心理咨询平台提供的服务是AI辅助心理咨询，后台由真人心理咨询师提供服务。

3: I understand that the service provided by the current counseling platform at this stage is AI-assisted psycho-counseling, and the service is provided by a real human counselor.

4: 我理解线上文字咨询服务为

4: I understand that the online text-based counseling service is: 

5: 互联网线上文字形式的即时心理困惑解答和心理知识普及服务。该服务为中文服务，且仅在此心理咨询平台上提供。

5: Instant psychological puzzle answering and psychological knowledge popularization services in the form of online text on the Internet. This service is provided in Chinese and is only available on the current counseling platform.

6: 我理解服务内容为：

6: I understand the service content is:

7: 围绕心理困惑（包括但不限于：情绪问题、情感问题、家庭关系、人际关系、个人成长、生涯发展等困惑）提供支持与帮助。虽然难以保证彻底改善心理状况和解答困惑，但此心理咨询平台秉持着“有时治愈，常常帮助，总是安慰”的态度竭诚为您服务

7: Provide support and help around psychological confusion (including but not limited to: emotional problems, family relationships, interpersonal relationships, personal growth, career development, etc.). Although it is difficult to ensure that the psychological condition is completely improved and the confusion is solved, the current counseling platform upholds the attitude of "sometimes heals, often helps, always comforts" and serves you wholeheartedly.

8: 我理解在咨询过程中：

8: I understand that during the counseling process:

9: 谈话将涉及我的生理/心理健康及情绪状态等相关信息。我在咨询服务中享有隐私权，我所透露的个人信息原则上能被严格保密。同时，我的隐私权在内容和范围上受到国家法律的保护和约束。

9: The conversation will involve my physical/mental health and emotional state and other related information. I have the right to privacy in the counseling service. In principle, the personal information I disclose can be kept strictly confidential. At the same time, my right to privacy is protected and restricted by the national laws.

10: 我理解，基于国家法律，保密原则有包括但不限于以下条目的例外情况：

10: I understand that based on the national laws, the confidentiality principle has exceptions including but not limited to the following items:

11: 当寻求服务者或其他人准备或正在实施危害自身或他人人身、财产安全的行为时；

11: When the service seeker or other person is preparing or is performing an act that endangers himself or others' personal and property safety;

12: 当寻求服务者有可能危及他人时（例如传染病等情况）；

12: When the service seeker is likely to endanger others (such as infectious diseases, etc.)

13: 当寻求服务者透露的信息涉及未成年人正在或即将受到性侵犯时；

13: When the information disclosed by the service seeker involves the minor beings or about to be sexually assaulted;

14: 当寻求服务者或其他人准备或正在实施危害国家安全、公共安全的行为时；

14: When the service seeker or other person is preparing or is performing an act that endangers national security or public security;

15: 在数据脱敏后实现匿名的情况下，当咨询团队成员之间讨论、请教或者接受督导和培训时；

15: When the data is desensitized and anonymous, when discussing, counseling, or receiving supervision and training among members of the counseling team;

16: 在数据脱敏后实现匿名的情况下，用于科学研究时；

16: When data is desensitized and anonymous, when used in scientific research;

17: 当法律规定需要披露时。

17: When the law requires disclosure.

18: 我同意对于以上非保密情况，出于保护我或相关人员的根本权利的原因，此心理咨询平台可仅在必要的人员范围内予以最小程度的信息披露。 此外，我理解，由于咨询服务基于网络进行，尽管此心理咨询平台致力于最大限度保护使用者的隐私，但是难以避免存在因网络信息安全漏洞、技术故障或未经授权的他人入侵等原因泄露使用者个人信息的可能性。

18: I agree that for the above non-confidentiality, for the purpose of protecting the fundamental rights of me or related personnel, the current counseling platform can only disclose information to a minimum within the scope of necessary personnel. In addition, I understand that since the counseling service is based on Online, although the current counseling platform is committed to protecting users’ privacy to the utmost extent, it is difficult to avoid the possibility of leaking users’ personal information due to network information security loopholes, technical failures, or unauthorized intrusions by others.

19: 我理解，当出现以下情况时，此心理咨询平台难以提供有效的咨询服务，需要寻求专业的线下治疗或咨询服务：

19: I understand that when the following situations occur, the current counseling platform is difficult to provide effective counseling services and needs to seek professional offline treatment or counseling services:

20: 有自杀的想法或计划；

20: Have suicidal thoughts or plans;

21: 有伤害自身或他人的想法或计划；

21: Have thoughts or plans to harm oneself or others;

22: 有经医院确诊的任一精神疾患；

22: Any mental illness diagnosed by the hospital;

23: 符合任一精神障碍诊断标准。

23: Meet any of the diagnostic criteria for mental disorders.

24: 我理解，如果我在信息中描述或体现出的生理、心理、精神状态以及行为计划符合以上任一标准的，此心理咨询平台无法继续为我提供咨询服务，且可能会建议我寻求专业的线下治疗或咨询服务。

24: I understand that if the physical, psychological, mental state and behavior plan described or reflected in the information meets any of the above standards, the current counseling platform cannot continue to provide me with counseling services and may suggest that I seek professional advice Offline treatment or counseling services.

25: 我理解，此心理咨询平台围绕心理困惑（包括但不限于：情绪问题、情感问题、家庭关系、人际关系、个人成长、生涯发展等困惑）提供支持与帮助，但仍然存在一些难以实现的服务：

25: I understand that the current counseling platform provides support and help around psychological confusion (including but not limited to: emotional problems, family relationships, interpersonal relationships, personal growth, career development, etc.), but there are still some services that are difficult to achieve:

26: 自杀或其他伤害行为的危机干预；

26: Crisis intervention for suicide or other harmful behaviors;

27: 精神障碍的诊断与治疗；

27: Diagnosis and treatment of mental disorders;

28: 精神类药物使用方面的具体建议；

28: Specific recommendations on the use of psychotropic drugs;

29: 处理重度心理创伤；

29: Deal with severe psychological trauma;

30: 为我提供具体的职业、学业等资源或信息；

30: Provide me with specific career, academic and other resources or information;

31: 为我提供对于社会现象的看法和政策解读；

31: Provide me with my views on social phenomena and policy interpretation;

32: 为我解梦（例如告诉我梦的含义、为什么会梦到什么人或事等）；

32: Interpret my dreams (for example, tell me the meaning of dreams, why I dreamed of people or things, etc.);

33: 为我解答非我本人的心理困惑(例如我的朋友、家人、网友等）。

33: Solve for me the psychological confusion that is not myself (such as my friends, family members, netizens, etc.).

34: 我理解，当我描述的情况超出此心理咨询平台的服务范围（服务范围不包括以上8种）时，此心理咨询平台无法满足我的咨询需求。

34: I understand that when the situation I describe exceeds the scope of the current counseling platform's services (the scope of services does not include the above 8 types), the current counseling platform cannot meet my counseling needs.↵"
37:

35: 我理解互联网线上文字咨询服务潜在的益处和风险

35: I understand the potential benefits and risks of Internet online text counseling services

36: 其中的益处包括但不限于：

36: The benefits include but are not limited to:

37: 能够更便捷地获取服务，无需前往约定的地点。并且，尽管风险很小，但我仍理解可能存在的风险。

37: You can get services more conveniently without going to the agreed place. And, although the risk is small, I still understand the possible risks.

38: 潜在风险包括但不限于：

38: Potential risks include but are not limited to:

39: 由于我提供的信息可能是不充分的，我得到的服务无法充分解答我的困惑或改善我的心理状态；由于可能存在的技术故障或其他不可预见的原因，我无法及时得到对我心理困惑的分析与建议。

39: Because the information I provided may be inadequate, the service I received cannot adequately answer my confusion or improve my mental state; due to possible technical failures or other unforeseen reasons, I cannot get timely information to me. Analysis and suggestions of psychological confusion.

40: 我同意，当为我提供咨询服务时，此心理咨询平台遵循的是中国大陆地区相关的法律法规，而非我所在地区的相关法律法规。

40: I agree that when providing counseling services for me, the current counseling platform follows the relevant laws and regulations in mainland China, not the relevant laws and regulations in my area.

41: 以上知情同意在我单次或多次使用服务的过程中持续生效.

41: The above informed consent will continue to take effect during my single or multiple uses of the service.

\end{CJK*}
\end{document}